# Exploration of carbonate aggregates in road construction using ultrasonic and artificial intelligence approaches

**Mohamed Abdelhedi**
mohamed.abdelhedi.etud@fss.usf.tn
Research Laboratory GEOMODEL (LR16ES17), Department of Earth Sciences, Faculty of Sciences, University of Sfax
Sfax, Tunisia

**Rateb Jabbar**
rateb.jabbar@qu.edu.qa

KINDI Center for Computing Research, College of Engineering, Qatar University, Doha, Qatar

**Chedly Abbes**
chedlyabbes8@gmail.com
Research Laboratory GEOMODEL (LR16ES17), Department of Earth Sciences, Faculty of Sciences, University of Sfax
Sfax, Tunisia

**ABSTRACT**

The COVID-19 pandemic has significantly impacted the construction sector, which is sensitive to economic cycles. In order to boost value and efficiency in this sector, the use of innovative exploration technologies such as ultrasonic and Artificial Intelligence techniques in building material research is becoming increasingly crucial. In this study, we developed two models for predicting the Los Angeles (LA) and Micro Deval (MDE) coefficients, two important geotechnical tests used to determine the quality of rock aggregates. These coefficients describe the resistance of aggregates to fragmentation and abrasion. The ultrasound velocity, porosity, and density of the rocks were determined and used as inputs to develop prediction models using multiple regression and an artificial neural network. These models may be used to assess the quality of rock aggregates at the exploration stage without the need for tedious laboratory analysis.

## 1 INTRODUCTION

Over the past decade, there has been a significant increase in the development of new methods for subsurface geological exploration, particularly in the exploration of building materials (Pell *et al.*, 2021). The increasing global economic crises have emphasized the need for new exploration methods such as the ultrasonic method and the application of artificial intelligence in the exploration of georesources, including building materials, particularly for developing countries. The construction of smart cities, such as the over 300 projects in China and the more than 100 planned in India, require a significant amount of building materials, particularly for road construction. Transportation is a fundamental function of a smart city (Toh *et al.*, 2020). Additionally, the population of the world's urban areas is increasing by 200,000 people per day, seeking affordable housing as well as social, transportation, and utility infrastructure. This has led to an increasing demand for sustainable construction materials globally (Marangu *et al.*, 2017).

The traditional methods for construction materials prospecting seem inadequate compared to the significant international demand for these materials. These techniques are labor-intensive, costly, and time-consuming. Therefore, finding more practical, faster, and less expensive prospecting techniques is of great interest (Abdelhedi and Abbes, 2021). While the ultrasonic method has not yet been well-developed for geomaterial applications, it is a very attractive tool. In recent years, several authors have applied this method in the exploration of carbonate rocks (Abdelhedi *et al.*, 2017), in gold mine exploration (De Souza *et al.*, 2022), and in mortar quality control (Abdelhedi *et al*, 2018).

Artificial intelligence (AI) is a set of computational algorithms used for clustering, predicting, and classifying tasks (Ebid, 2020). Because AI has superhuman abilities, it has revolutionized all spheres of technology and science (Jabbar, Jabbar and Kamoun, 2022; Moulahi *et al.*, 2022). AI is becoming more ubiquitous across several areas, including healthcare (Elleuch *et al.*, 2021), agriculture (Ayadi*et al.*, 2020), sustainability (Jabbar *et al.*, 2021; Abulibdeh, Zaidan and Jabbar, 2022; Zaidan *et al.*, 2022) and transportation (Jabbar *et al.*, 2018; Ben Said and Erradi, 2022).

In the field of geology, AI technology has attracted significant academic and industrial attention in recent years. AI has been applied in geoscience for the determination of reservoir rock properties, drilling optimization, and enhanced production facilities (Solanki *et al.*, 2022). Furthermore, these techniques have been used in carbonate rock exploration for the prediction of rocks and mortar UCS (Uniaxial Compressive Strength) values (Abdelhedi *et al.*, 2020). Additionally, AI has been applied in mining and geological engineering, including rock mechanics, mining method selection, mining equipment, drilling-blasting, slope stability, and environmental issues (Bui, Bui and Nguyen, 2021).

This study applied AI and ultrasonic methods to establish predictive models linking porosity, density, and ultrasonic velocity to Los Angeles (LA) and Micro Deval (MDE) coefficients, with the aim of increasing the exploration of high-quality carbonate aggregates used in road construction.

The remainder of this paper is organized as follows. Section 2 presents the methods and materials used in this study. Section 3 discusses the computational results obtained from experiments. Finally, in Section 4, conclusions are presented.

## 2   METHODS AND MATERILS

Seven samples were collected from carbonate formations and crushed into aggregates with a particle size ranging from 10 to 14 mm.

### 2.1   Ultrasonic velocity

The ultrasonic method involves exciting the structure of the material using a vibratory source, such as a piezoelectric transducer, which converts electrical energy into mechanical energy and vice versa. The use of transducers allows the control over the shape and duration of the pulse, providing a repetitive and energetic source. The measurements were carried out with direct contact, requiring the use of a coupling material between the transducer and the sample to reduce the loss of the signal (Abdelhedi et al. 2018).

### 2.2   Artificial Neural Network ANN

Artificial neural networks (ANNs) are a type of AI algorithm that is used to solve complex nonlinear problems. ANNs are modelled after the structure of the human brain and are composed of interconnected nodes, or "neurons," that process and transmit information. ANNs have been applied as a method of artificial intelligence in a variety of fields, including geomaterials applications.

In this study, we used an Artificial neural network (ANN) with three layers: an input layer, an output layer, and a hidden layer. The input layer is composed of three neurons, each representing a different variable (ultrasonic pulse velocity, density, and effective porosity) that is used as input data for the ANN. The output layer is formed by a single output neuron, which produces the predicted value of the LA or MDE coefficient based on the input data. The hidden layer is a single layer that processes the input data and generates an intermediate output that is used by the output layer to produce the final result.

The back-propagation neural network (BP-NN) (Wengang *et al.*, 2019) used in this study is a learning algorithm that is used to train the ANN. The BP-NN uses a technique called "back-propagation" to adjust the weights of the connections between the neurons in the ANN based on the difference between the predicted output and the actual output. The BP-NN adjusts the weights during a number of iterations, called "epochs," and uses a learning function called Levenberg-Marquardt (TRAINLM) (Kipli *et al.*, 2012) to determine the rate at which the weights are adjusted. By iteratively adjusting the weights in this way, the BP-NN is able to "learn" from the input data and improve the accuracy of its predictions over time.

## 2.3 Los Angeles coefficient

The resistance to fragmentation of aggregates is an important characteristic that can affect the performance of materials in various applications, such as road construction. To determine this property, the Los Angeles (LA) coefficient is commonly used. The LA coefficient is measured according to standards P18-572 (1990) (Amrani *et al.*, 2019) using a test that involves subjecting the material to standard ball shocks in a Los Angeles machine. The mass of the ball load used in the test varies based on the granular class of the material. The resistance to fragmentation is calculated as:

$$LA = 100 \frac{m}{M} \quad (1)$$

Where M is the mass of the material being tested and m is the mass of the particles smaller than 1.6 mm produced during the test.

## 2.4 Micro-Deval coefficient

The abrasion resistance of aggregates is an important property that can impact their performance in various applications. To determine this characteristic, the micro-deval coefficient is commonly used. This test, described in standards P18-573 (1990), measures the wear resistance of rocks under both dry and wet conditions. The test involves subjecting the material to reciprocal friction in a rotating cylinder under controlled conditions, using abrasive filler for tests on gravel between 10 and 14 mm. The micro-deval coefficient (MDE) in the presence of water is calculated as

$$MDE = 100 \frac{m}{M} \quad (2)$$

Where M is the mass of the material being tested and m is the mass of the particles smaller than 1.6 mm produced during the test. The wear resistance of the material is expressed by the quantity of these particles

## 3 RESULTS AND DISCUSSION

Artificial neural networks (ANNs) have been shown to be effective models for predicting complex rock properties in multiple studies (Kahraman *et al.*, 2010; Madhubabu *et al.*, 2016).

Therefore, ANNs were used to create two models that linked the physical parameters to the mechanical parameters of carbonate rock aggregates, allowing for the estimation of their resistance to fragmentation and abrasion. Ultrasound velocity, porosity, and density were first determined and analyzed, and then used to create predictive models using Excel and ANN. The data were used for training, validation, and verification of the prediction efficiency of the models. The input parameters for the models were ultrasonic velocity, porosity, and density, while the outputs were either the MDE or LA coefficient. The MDE and LA coefficient values predicted by the models created using multiple regression are plotted against the measured values in Figures 1 and 2, respectively.

We note that these two relationships exhibited significant correlations ($R^2 > 0.8$). However, the first correlation shows a negative estimated value of MDE, which is not acceptable.

The models produced by the artificial neural network give two correlations between the predicted MDE and LA coefficient values and the measured values (figures 3 and 4 respectively). The plotted data for the output parameters for both models were close to Line 1, indicating a favourable prediction. Models generated by ANN analysis were more accurate than models generated by multiple regressions. Figure 1 shows that the multiple regression model produced a false estimation of MDE (a negative value), demonstrating the ANN model's superiority.

Abdelhedi et al. (Abdelhedi *et al.*, 2020) conducted a similar study in which they used ANNs to predict uniaxial compressive strength (UCS) in carbonate rocks. Tariq et al. (Tariq *et al.*, 2017) used ultrasonic pulse velocity and density to develop an ANN model with a correlation coefficient of $R^2 = 0.84$ to predict UCS values in carbonate rocks.

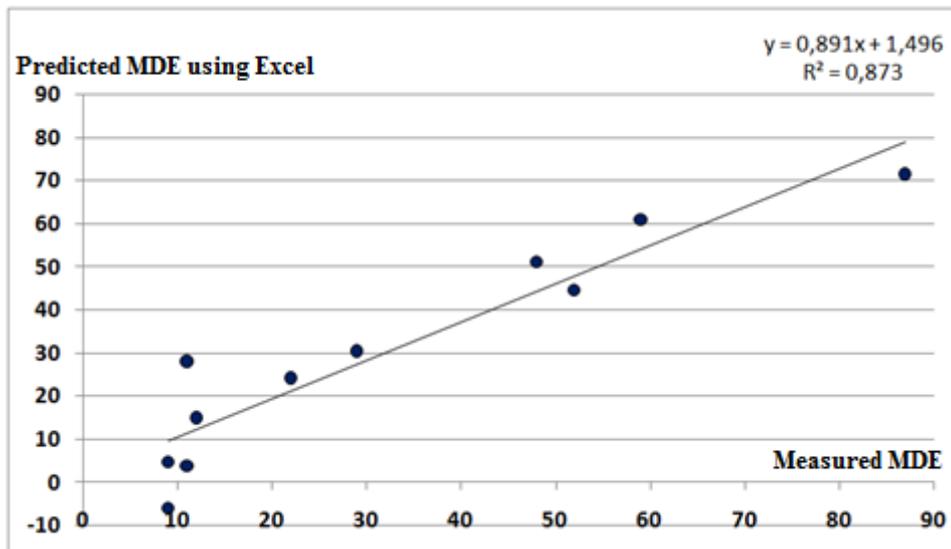

Figure 1: Correlation between MDE values estimated by multiple regression and measured MDE

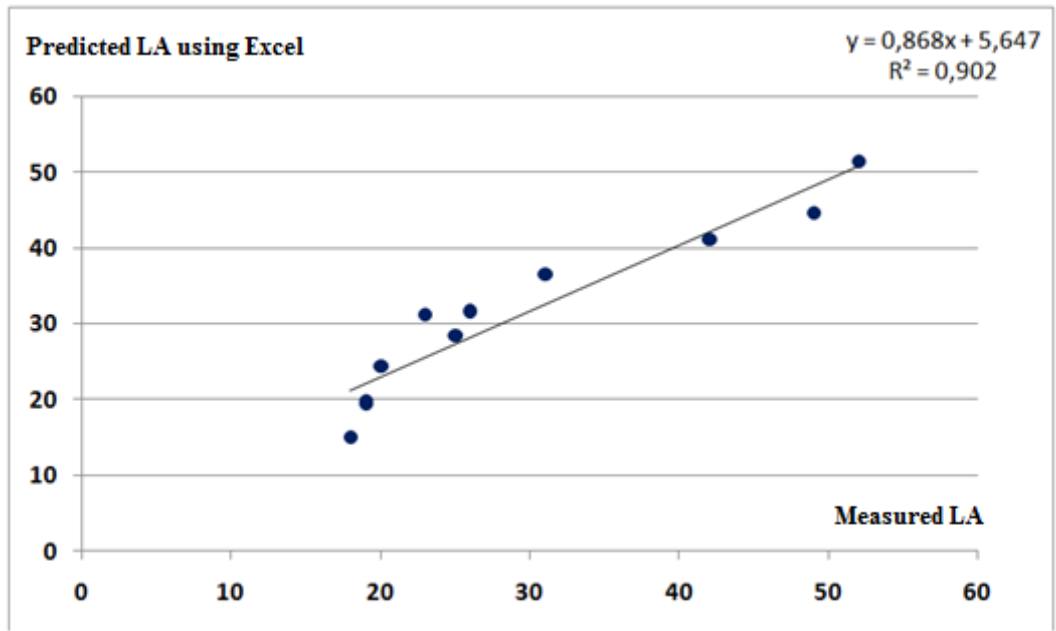

Figure 2: Correlation between LA values estimated by multiple regression and measured LA

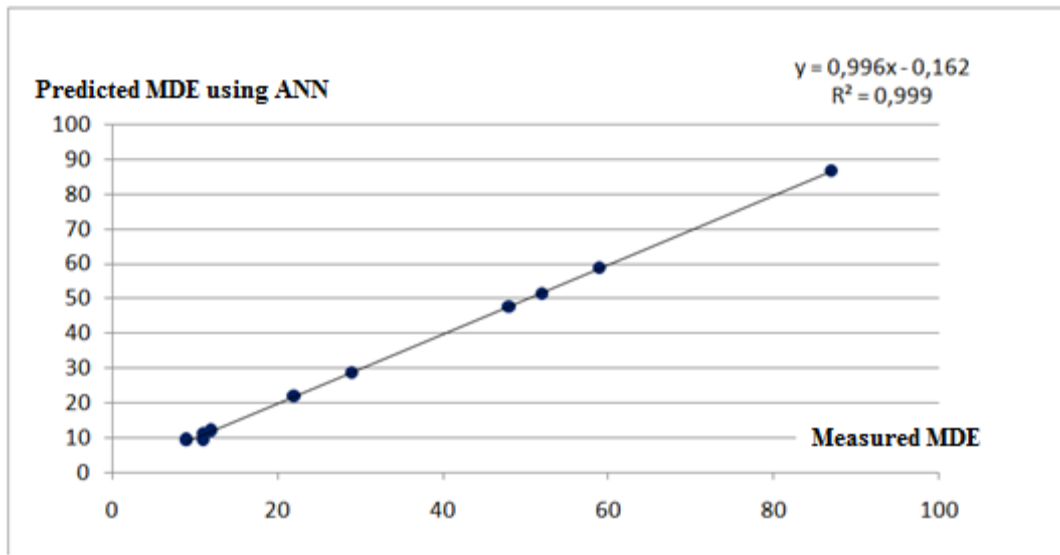

Figure 3: Correlation between MDE values estimated by ANN and measured MDE

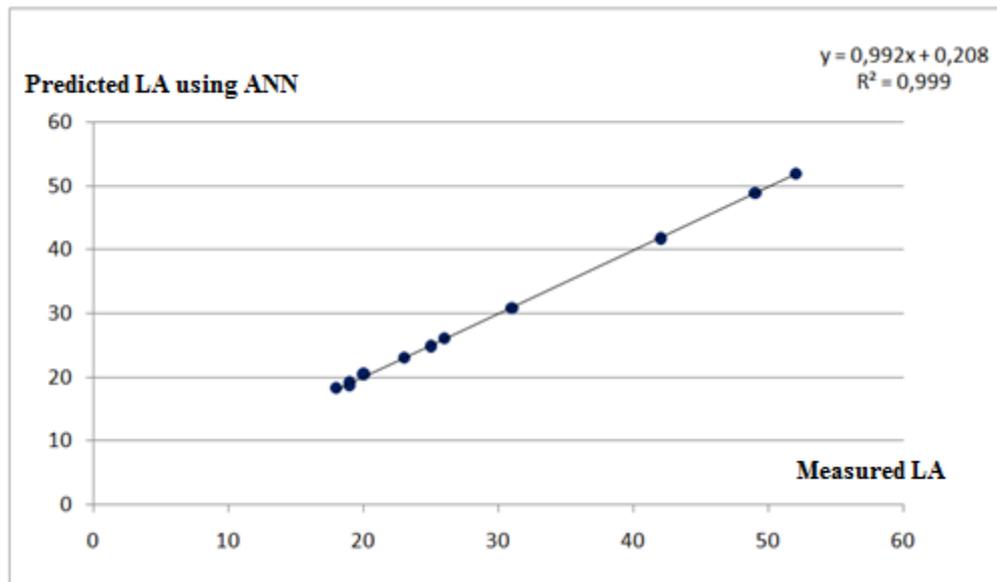

Figure 4: Correlation between LA values estimated by ANN and measured LA

## 4 CONCLUSION

The performance of construction materials, such as aggregates, can significantly affect the strength and durability of structures. In order to evaluate the suitability of different aggregates for different applications, it is important to measure their physical and mechanical properties. Two important properties that are commonly tested for aggregates are their resistance to fragmentation and abrasion. These properties can be evaluated using the Los Angeles (LA) and micro-deval (MDE) tests, respectively.

In this study, a number of carbonate rock aggregates were subjected to the LA and MDE tests in order to determine their resistance to fragmentation and abrasion. Multiple correlation methods were then used to develop predictive models linking the physical parameters of porosity, density, and ultrasonic velocity to the LA and MDE coefficients. These models can be used to predict the resistance to fragmentation and abrasion in the exploration of new mining sites of aggregates based on their physical properties.

The LA and MDE coefficients are particularly important for aggregates used in the construction of roads and hydraulic concretes. By using this technique, it is possible to easily assess the potential of new aggregate georesources for use in these applications.

Despite some limitations, such as the small sample size, this research is an important step towards improving our understanding of the behavior of aggregates and their suitability for use in construction.

# REFERENCES

Abdelhedi, Mohamed, Aloui, Monia, Mnif, Thameur, & Abbes, Chedly, (2017). "Ultrasonic velocity as a tool for mechanical and physical parameters prediction within carbonate rocks", Geomechanics and Engineering, 13(3), 371-384. DOI: https://doi.org/10.12989/gae.2017.13.3.371

Abdelhedi, Mohamed, Mnif, Thameur, & Abbes, Chedly (2018). "Ultrasonic velocity as a tool for physical and mechanical parameters prediction within geo-materials: Application on cement mortar", Russian Journal of Nondestructive Testing, 54(5), 345-355. DOI: https://doi.org/10.1134/S1061830918050091

Abdelhedi, Mohamed, Jabbar, Rateb, Mnif, Thameur, & Abbes, Chedly, (2020). "Prediction of uniaxial compressive strength of carbonate rocks and cement mortar using artificial neural network and multiple linear regressions", Acta Geodynamica et Geromaterialia, 17(3), 367-378. DOI: 10.13168/AGG.2020.0027

Abdelhedi, Mohamed, & Abbes, Chedly, (2021). "Study of physical and mechanical properties of carbonate rocks and their applications on georesources exploration in Tunisia". Carbonates and Evaporites, 36(2), 1-13. https://doi.org/10.1007/s13146-021-00688-8

Abulibdeh, Ammar, Zaidan, Esmat. & Jabbar, Rateb, (2022). "The impact of COVID-19 pandemic on electricity consumption and electricity demand forecasting accuracy: Empirical evidence from the state of Qatar", Energy Strategy Reviews, 44, p. 100980. https://doi.org/10.1016/j.esr.2022.100980

Amrani, Mustapha, Taha, Yassine, Kchikach, Azzouz, Benzaazoua, Mostafa, & Hakkou, Rachid, (2019). "Valorization of phosphate mine waste rocks as materials for road construction", Minerals, 9(4), 237. https://doi.org/10.3390/min9040237

Ayadi, Safa, Ben Said, Ahmed, Jabbar, Rateb, Aloulou, Chafik, Chabbouh, Achraf, & Ben Achballah, Ahmed, (2020). "Dairy cow rumination detection: A deep learning approach", In International Workshop on Distributed Computing for Emerging Smart Networks, 123-139. Springer, Cham. DOI: https://doi.org/10.1007/978-3-030-65810-6_7

Bui, Xuan-Nam, Bui, Hoang-Bac, & Nguyen, Hoang, (2021). "A Review of Artificial Intelligence Applications in Mining and Geological Engineering", Proceedings of the International Conference on Innovations for Sustainable and Responsible Mining, 109, pp. 109–142. DOI: https://doi.org/10.1007/978-3-030-60839-2_7

De Souza, Andre Eduardo Calazans Matos, Sarout, Joel, Luzin, Vladimir, Sari, Mustafa, & Vialle, Stefanie, (2022). "Laboratory measurements of ultrasonic wave velocities and anisotropy across a gold-hosting structure: A case study of the Thunderbox Gold Mine, Western Australia", Ore Geology Reviews, 146, 104928. https://doi.org/10.1016/j.oregeorev.2022.104928

Ebid, Ahmed-M, (2020). "35 Years of (AI) in Geotechnical Engineering: State of the Art", Geotechnical and Geological Engineering, 2, 39(2), pp. 637–690. DOI: https://doi.org/10.1007/s10706-020-01536-7

Elleuch, Mohamed-Ali, Ben Hassena, Amal, Abdelhedi, Mohamed, & Francisco Silva, Pinto, (2021). "Real-time prediction of COVID-19 patients health situations using Artificial Neural Networks and Fuzzy Interval Mathematical modeling", Applied Soft Computing, 110, p. 107643. https://doi.org/10.1016/j.asoc.2021.107643

Jabbar, Rateb, Zaidan, Esmat, ben Said, Ahmed, & Ghofrani, Ali, (2021). "Reshaping Smart Energy Transition: An analysis of human-building interactions in Qatar Using Machine Learning Techniques", arXiv preprint arXiv:2111.08333. https://doi.org/10.48550/arXiv.2111.08333

Jabbar, Rateb, Khalifa, Al-Khalifa, Mohamed, Kharbeche, Wael, Alhajyaseen, Mohsen, Jafari, & Shan, Jiang (2018). "Applied Internet of Things IoT: Car monitoring system for Modeling of Road Safety and Traffic System in the State of Qatar", Qatar Foundation Annual Research Conference Proceedings Volume 2018 Issue 3. Vol. 2018. No. 3. Hamad bin Khalifa University Press (HBKU Press). https://doi.org/10.5339/qfarc.2018.ICTPP1072

Jabbar, Rahma, Jabbar, Rateb, & Kamoun, Slaheddine (2022). "Recent progress in generative adversarial networks applied to inversely designing inorganic materials: A brief review." Computational Materials Science 213 (2022): 111612. https://doi.org/10.1016/j.commatsci.2022.111612

Kahraman, Sair, Alber, Michael, Fener, Mustafa, & Gunaydin, Osman, (2010)."The usability of Cerchar abrasivity index for the prediction of UCS and E of Misis Fault Breccia: regression and artificial neural networks analysis". Expert Systems with Applications, 37(12), 8750-8756. https://doi.org/10.1016/j.eswa.2010.06.039

Kipli, Kuryati, Muhammad, Mohd Saufee, Masra, Sh. Masniah Wan, Zamhari, Nurdiani, Lias, Kasumawati, & Mat, Dayang Azra Awang, (2012). "Performance of Levenberg-Marquardt backpropagation for full reference hybrid image quality metrics", Proceedings of International Conference of Muti-Conference of Engineers and Computer Scientists (IMECS'12) (pp. 704-707).

Madhubabu, N, Singh, PK, Kainthola, Ashutosh, Mahanta, Bankim, Tripathy, A, & Singh, Tn, (2016). "Prediction of compressive strength and elastic modulus of carbonate rocks", Measurement, 88:202-213. https://doi.org/10.1016/j.measurement.2016.03.050

Marangu, Joseph Mwiti, Latif, Eshrar, and Maddalena, Riccardo, (2021). "Evaluation of the reactivity of selected rice husk ash-calcined clay mixtures for sustainable cement production." Edited by R. Maddalena and M. Wright-Syed: 81.

Moulahi, Tarek, Jabbar, Rateb, Alabdulatif, Abdulatif, Abbas, Sidra, El Khediri, Salim, Zidi, Salah, & Rizwan, Muhammad, (2022). "Privacy-preserving federated learning cyber-threat detection for intelligent transport systems with blockchain-based security". Expert Systems, e13103. https://doi.org/10.1111/exsy.13103

Pell, Robert, Tijsseling, Laurens, Goodenough, Kathryn, Wall, Frances, Dehaine, Quentin, Grant, Alex, Deak, David, Yan, Xiaoyu, ... & Whattoff, Phoebe, (2021). "Towards sustainable extraction of technology materials through integrated approaches", Nature Reviews Earth & Environment, 2(10), 665-679. https://doi.org/10.1038/s43017-021-00211-6

Ben Said, Ahmed. & Erradi, Abdelkarim, (2022) "Spatiotemporal Tensor Completion for Improved Urban Traffic Imputation", IEEE Transactions on Intelligent Transportation Systems, 23(7), pp. 6836–6849. DOI: 10.1109/TITS.2021.3062999

Solanki, Parth, Baldaniya, Dhruv, Jogani, Dhruvikkumar, Chaudhary, Bhavesh, Shah, Manan & Kshirsagar, Ameya, (2022). "Artificial intelligence: New age of transformation in petroleum upstream", Petroleum Research, 7(1), pp. 106–114. https://doi.org/10.1016/j.ptlrs.2021.07.002

Standard P18-572 (1990) Aggregates. Micro-deval attribution test. Association française de normalisation, Bureau de Normalisation Sols et Routes. Géotechnique-normes. Décembre 1990, Paris, France.

Standard P18-573 (1990) Aggregates. Los Angeles test–granulate. Los Angeles pruefung. Association française de normalisation, Bureau de Normalisation Sols et Routes. Géotechnique-normes. Décembre 1990, Paris, France.

Tariq, Zeeshan, Elkatatny, Salaheldin, Mahmoud, Mohammed., Ali, Abdelwahab Z, & Abdulraheem, Abdulazeez, (2017). "A new technique to develop rock strength correlation using artificial intelligence tools", SPE Middle East Oil and Gas Show and Conference, 18-21, March, Manama, Bahrain. https://doi.org/10.2118/186062-MS

Toh, Chai K, Sanguesa, Julio A, Cano, Juan C, & Martinez, Francisco J, (2020). "Advances in smart roads for future smart cities", Proceedings of the Royal Society A, 476(2233), 20190439. https://doi.org/10.1098/rspa.2019.0439

Wengang, Zhang, Goh, AnthonyTeck Chee, Runhong, Zhang, Yongqin, Li, & Ning, wei, (2020). "Back-propagation neural network modeling on the load–settlement response of single piles." Handbook of probabilistic models. Butterworth-Heinemann, 2020. 467-487. https://doi.org/10.1016/B978-0-12-816514-0.00019-9

Zaidan, Esmat, Abulibdeh, Ammar., Alban, Ahmed, & Jabbar, Rateb, (2022). "Motivation, preference, socioeconomic, and building features: New paradigm of analyzing electricity consumption in residential buildings", Building and Environment, 109177. https://doi.org/10.1016/j.buildenv.2022.109177